\documentclass[11pt]{article}

\usepackage[final]{acl}
\usepackage{times}
\usepackage{latexsym}
\usepackage[T1]{fontenc}
\usepackage[utf8]{inputenc}
\usepackage{microtype}
\usepackage{algorithm}
\usepackage{algpseudocode}
\usepackage{float}
\usepackage{hyperref}
\usepackage{inconsolata}
\usepackage{graphicx}

\title{AlignCheck: a Semantic Open-Domain Metric for Factual Consistency Assessment}

\author{Ahmad Aghaebrahimian \\
Institute of Computational Life Sciences,\\
Department of Life Sciences and Facility Management,\\
Zurich University of Applied Sciences, 8820 Waedenswil, Switzerland\\ 
Swiss Institute of Bioinformatics, 1015 Lausanne, Switzerland \\
  \texttt{ahmad.aghaebrahimian@zhaw.ch}}

\begin{document}
\maketitle
\begin{abstract}
Large Language Models have significantly advanced natural language processing tasks, but remain prone to generating incorrect or misleading but plausible arguments. This issue, known as hallucination, is particularly concerning in high-stakes domains like clinical applications, where factual inaccuracies can have severe consequences. Existing evaluation metrics fail to adequately assess factual consistency and lack interpretability, making diagnosing and mitigating errors difficult. We propose an interpretable framework for factual consistency assessment for in-domain and open-domain texts to address these limitations. Our approach decomposes text into atomic facts and introduces a flexible, schema-free methodology. Unlike previous methods with an absolute metric, we incorporate a weighted metric to enhance factual evaluation. Additionally, we propose a mechanism to control assessment complexity in intricate domains. We benchmark our approach on popular general and clinical datasets and release our code to support fact-aware model training in future research.
\end{abstract}

\section{Introduction}
\label{sec:SCIENTIFIC-BACKGROUND}
Large Language Models (LLMs) have revolutionized various natural language generation tasks, including question answering~\cite{deutsch-etal-2021-towards}, text summarization~\cite{goyal2023newssummarizationevaluationera}, and dialogue systems~\cite{shuster-etal-2021-retrieval-augmentation}. Despite their impressive capabilities, LLMs are prone to a phenomenon known as hallucination, where they generate incorrect or misleading arguments with high confidence. This issue is particularly critical in high-stakes domains such as clinical and medical applications, where factual inaccuracies can have severe consequences. 

To mitigate these risks, it is mandatory to validate the factual consistency of LLM-generated content. Although numerous methods have been proposed to assess factual accuracy~\cite{Lee2022factuality,min-etal-2023-factscore, Goodrich2019}, many suffer from key limitations. Specifically, existing approaches often lack interpretability, offering only numerical~\cite{deutsch-etal-2021-towards} or binary~\cite{tang-etal-2023-understanding} metrics without indicating where within the generated text errors occur. This absence of granular insight makes it difficult to diagnose and correct inaccuracies effectively. As highlighted by Luo et al.~\cite{luo2025factualconsistencyevaluationsummarization}, current methods for factual consistency checking fall short, particularly for clinical texts. Additionally, current techniques do not provide sufficient flexibility to account for different classes of facts, such as facts about a patient's demography or health journey, limiting their applicability in diverse real-world scenarios. This highlights the need for more robust, interpretable, and adaptable evaluation methods for factual consistency checking that align better with human judgments and task-specific requirements. 

In this paper, we propose an interpretable framework for factual consistency checking on general and clinical texts, showcasing the application of the framework for summarization. Similar to earlier studies~\cite{goyal2023newssummarizationevaluationera}, we advocate decomposing texts into atomic facts for more granular analysis. However, in contrast to most methods, which focus on the sentence-level fact assessment~\cite{goyal2023newssummarizationevaluationera}, we adopt a general approach to broaden the scope of consistency checking to the global text, similar to Min et al~\cite{min-etal-2023-factscore}. In contrast to Min et al.~\cite{min-etal-2023-factscore}, though, our approach is schema-free and does not depend on an external knowledge base or schema. To accommodate schema-free factuality evaluation, we introduce a weighted metric inspired by the F1 score, combined with BERTScore~\cite{zhang2020bertscoreevaluatingtextgeneration}. Furthermore, we suggest a flexible mechanism to control the complexity of factual evaluation when dealing with intricate domains.

We release the code\footnote{\url{https://github.com/Soshaince/AlignCheck}} and suggest integrating the score into the objective function for training a fact-aware model in further studies.

In summary, our contribution consists of
\begin{itemize}
    \item AlignCheck, a new F1 score formulation for factual consistency, complemented with the codebase,
    \item a granular algorithm for different levels of fact-checking,
    \item and benchmarking the factuality score for popular factual datasets.
\end{itemize}

\section{Related Work}
Evaluating the quality of synthetic text, such as the output of machine translation or summarization algorithms, has always been an issue due to the inherent subjectivity and complexity of language. Popular metrics such as BLEU (Bilingual Evaluation Understudy)~\cite{papineni-etal-2002-bleu} and ROUGE (Recall Oriented Understudy for Gisting Evaluation)~\cite{lin-2004-rouge} focus on surface-level information, such as precision and recall over n-gram overlap, thus fail to capture deeper semantic fluency and coherence in the generated text. BERTScore~\cite{zhang2020bertscoreevaluatingtextgeneration} suggested a semantic score to address these shortcomings. However, BERTScore and other semantic-based metrics do not account for factual accuracy, meaning highly coherent text can still correspond to factually inconsistent outputs~\cite{li-etal-2024-leveraging-large,zhou-etal-2022-deconstructing}. 

According to recent studies~\cite{Goodrich2019,falke-etal-2019-ranking}, up to 30\% of synthetic summaries contain factual consistency problems. Therefore, improving factual consistency checking has been an active field in recent years~\cite{nori2023capabilitiesgpt4medicalchallenge,shuster-etal-2021-retrieval-augmentation}. Several studies suggested model-based factuality evaluation where a model is trained to evaluate the consistency for generated text by formulating it as a binary classification task~\cite{kryscinski-etal-2020-evaluating}, question answering~\cite{deutsch-etal-2021-towards}, or Natural Language Inference (NLI)~\cite{laban-etal-2022-summac} through textual entailment assessment. 

Decomposing text into constituents such as entities~\cite{Lee2022factuality} or triples (subject, predicate, object) is exercised in several other works~\cite{Goodrich2019,thorne-etal-2018-fever, min-etal-2023-factscore} as a direct and model-free alternative to validate factual consistency of generated texts. Our approach fits in this category and is similar to~\cite{Goodrich2019}; however, we do not assume a fixed schema for the task. Instead, we adopt a flexible approach by relaxing this assumption, thus broadening the domain of the application. 

\section{Methods and Data}
\label{sec:DATA-AND-METHODS}
We run a basic Named Entity Recognition (NER) tagger over the source (i.e., ground truth summary) and target (i.e., predicted summary) texts to get the sets of source $E_s$ and target entities $E_t$. We utilized two datasets, AgreeFact~\cite{tang-etal-2023-understanding} and MIMIC-IV-Ext-BHC~\cite{Aali_mimic2024}. The details of the datasets are provided below. For general texts in AgreeFact and clinical texts in MIMIC-IV-Ext-BHC, we used Spacy and MedCat, respectively. The NER tagger extracts named entities with their types from both source and target texts. In general domains, these types might be Person, Time, Organization, etc, and in the clinical context, they might be Diagnosis, Prognosis, Treatment, etc. We treat the output of NER annotators as a bag of types, thus assigning TD-IDF (Term Frequency-Inverse Document Frequency) weights to each type. Depending on the granularity of the intended factual assessment, we can decide how many top-n types $T$ we want to include in the assessment.

Given each source text, we define $F_s$ as the set of all facts $f_s = (s_s, p_s, o_s)$ where $s_s$ is a subject named entity $s_s \in E_s$ whose type $s_t \in T$ and which is associated with $o_s$, a type-free object named entity by the schema-less predicate $p_s$. Likewise, we define $F_t$ as the set of facts$f_t = (s_t, p_t, o_t), s_t \in E_t$ extracted from each target text. We also define $SO_s$ and $SO_t$ as a set of unique $(s_s, o_s)$ and $(s_t, o_t)$ accordingly. Given $F_s$ and $F_t$ as sets of triples, we can define True Positive(TP) as shared triples between $F_s$ and $F_t$, False Negatives(FN) as $F_s - (F_s \cap F_t)$ and False Positive(FP) as $F_t - (F_s \cap F_t)$. 

FP and FN both negatively impact the factual consistency. While an increase in the number of FNs reduces Recall(R) and signals factual inconsistency, an increase in the number of FPs reduces Precision(P) and might signal hallucination.

The F1 score, $F1=\frac{2PR}{P+R}$, as the harmonic mean of precision and recall, is a popular metric in many experiments with imbalanced labels. In schema-based systems~\cite{Goodrich2019}, it is easy to treat $F_s$ and $F_t$ as atomic facts with constant named entities and predicates, thus computing the F1 score as described above is straightforward. In schema-less systems, though, there might be inconsistency in predicates, thus, a fact with the same semantics, but not matching the predicate in $F_s$ will be considered as FN or FP. To address this issue, we use BERTScore as a means of soft similarity estimation. BERTScore computes the similarity between two strings as the sum of the cosine similarities between their token embeddings, thereby enabling paraphrase detection.

\begin{algorithm}[H]
\small
\caption{Weighted scoring algorithm} 
\label{alg}
\begin{algorithmic}
\State $TP \gets 0$
\State $FP \gets 0$
\State $FN \gets 0$
    \For{$f_s \in F_s$}
        \If{$(s_s, o_s) \notin SO_t$}
            \State $FN \gets FN + 1$
        \Else
            \For{$f_t \in F_t$}
                \If{$s_s == s_t$}
                    \If{$o_s == o_t$}
                        \State $TP \gets TP + BERTScore(p_s, p_t)$
                    \EndIf
                \EndIf
            \EndFor
        \EndIf
    \EndFor
    \For{$f_t \in F_t$}
        \If{$(s_t, o_t) \notin SO_s$}
            \State $FP \gets FP + 1$
        \EndIf
    \EndFor    
\end{algorithmic}
\end{algorithm}

As indicated in the Algorithm~\autoref{alg}, every missing $(s_s, o_s)$ in $SO_t$ is a full instance of FN. Conversely, each lacking instance of $(s_t, o_t)$ in $SO_s$ increases FP by one full unit. TP is impacted when $s_s == s_t$ and $o_s == o_t$. In this case, TP increments or decrements based on $BERTScore(p_s, p_t)$, which is in the range $(-1,1)$. This means that as the target predicate becomes semantically closer to the source predicate, the value of TP increases and decreases as the semantic distance grows.

To showcase the application of our proposed metric, we selected two datasets, AgreeFact~\cite{tang-etal-2023-understanding} in general and MIMIC-IV-Ext-BHC~\cite{Aali_mimic2024} in the clinical domain.

AgreeFact~\cite{tang-etal-2023-understanding} is a benchmark for assessing factual consistency as a binary classification. It consists of 9 annotated factuality datasets stratified according to the deployed summarization model. We employ samples of the dataset, consisting of 2353 texts, summarized with one or more of four different summarization models, including BART~\cite{lewis-etal-2020-bart}, Pegasus and PegasusDynamic~\cite{Zhang2020pegasus}, and T5~\cite{Raffel2020}. The dataset consists of 1726 unique samples. Since we intend to compare the factuality of different models, we ignored 1313 samples that were summarized using only one model.

MIMIC-IV-Ext-BHC~\cite{Aali_mimic2024} is a dataset of Brief Hospital Course (BHC) summaries. BHCs are clinical documents that summarize a patient’s hospital stay. MIMIC-IV-Ext-BHC is extracted from MIMIC-IV-Note~\cite{Johnson2023}, which is a raw collection of 331,794 de-identified discharge summaries from 145,915 patients admitted to the Beth Israel Deaconess Medical Center. To demonstrate the application of our proposed metric, we sampled 10000 BHCs for fine-tuning and 300 BHCs for testing. We finetuned an instance of Llama 3-13B with different strategies as described below.

The LM\_ep2 strategy involves post-tuning the LLM on 10,000 BHCs plain texts for two epochs, while LM\_ep5 extends this process to five epochs. In contrast, the Instruction\_ep2 strategy uses instruct-tuning, where the LLM is trained on 10,000 BHCs along with their summaries for two epochs, and Instruction\_ep5 increases the number of epochs for this instruct-tuning process to five. These strategies highlight different approaches to fine-tuning, varying both in training data usage and the number of epochs. 

Employing each model, we summarized the 300 test samples using the fine-tuned models.

\section{Results}
\label{sec:RESULTS}
We employed the AlignCheck to estimate the factual overlap between each summarization model's prediction and the ground-truth summary text.~\autoref{table_agree} summarizes the number of times each model performs the best with respect to the others on the AgreeFact and MIMIC-IV-Ext-BHC datasets.

\begin{table*}[t]
\centering
\begin{tabular}{ l l l l l}
    \bf Models              &\bf BART       &\bf Pegasus        &\bf PegasusDynamic  &\bf T5 \\ 
     \hline
    \bf BART                &1.0000         &1.110223e-16       &0.0000             &0.0000 \\
    \bf Pegasus             &1.110223e-16   &1.0000             &0.0000             &1.566292e-11 \\
    \bf PegasusDynamic      &0.0000         &0.0000             &1.0000             &0.0000 \\
    \bf T5                  &0.0000         &1.566292e-11       &0.0000             &1.0000 \\
    \hline
    \hline
    \bf Models          &\bf instruction\_ep2        &\bf instruction\_ep5        &\bf lm\_ep2         &\bf lm\_ep5\\ 
     \hline
    \bf instruction\_ep2     &1.0                        &8.728525e-01               &9.882236e-06         &0.001026 \\
    \bf instruction\_ep5     &0.872852                   &1.0                        &1.712586e-07         &0.000039 \\
    \bf lm\_ep2              &0.000010                   &1.712586e-07               &1.0                  &0.723773 \\
    \bf lm\_ep5              &0.001026                   &3.905937e-05               &7.237732e-01         &1.0 \\
    \hline
\end{tabular}
\caption{First table: Friedman posthoc test on AgreeFact~\cite{tang-etal-2023-understanding}. Second table: Friedman post hoc test on MIMIC-IV-Ext-BHC~\cite{Aali_mimic2024}. p-values lower than 0.05 show a statistically significant difference.}
\label{Friedman}
\end{table*}

\begin{table}
\centering
\begin{tabular}{ l l }
 \hline
 \bf Model (AgreeFact) & \bf First rank \\ 
  \hline
    BART & 93.7\% \\
    Pegasus & 3.6\% \\
    PegasusDynamic & 2.3\% \\
    T5 & 0.4\% \\ 
    \hline
    \bf Model (MIMIC-IV-Ext-BHC) & \bf First rank\\
    \hline
    instruction\_ep2 & 23.7\%\\
    instruction\_ep5 & 26.3\%\\
    lm\_ep2 & 37.4\%\\
    lm\_ep5 & 12.6\%\\ 
    \hline
\end{tabular}
\caption{Top ranks. Number of times each model gets the highest score on each dataset}
\label{table_agree}
\end{table}

Then we used the Friedman test to check if there is a difference between models characterized by the AlignCheck score. The Friedman test is a non-parametric statistical test used to detect differences in performance across more than 3 models evaluated on multiple data points, where each sample has a score for each model, and there is no normality assumption. The p-values $p=0.000$ estimated by the Friedman test for both datasets demonstrate that the models are statistically significantly different. The Friedman test tells if there’s a difference, but does not say which models differ. After the Friedman test, a follow-up with post-hoc tests reported in~\autoref{Friedman} shows which models are statistically significantly different as expected by their number of parameters and training protocols.

\section{\bf Conclusion and Future Work}
\label{sec:CONCLUSIONS}
In high-stakes domains such as clinical contexts, preserving critical facts, particularly of certain types, is essential when doing text processing tasks such as summarization. In this study, we introduced a novel schema-free methodology and scoring algorithm for assessing factual consistency in both in-domain and open-domain contexts. Our approach involves decomposing source and target texts into atomic facts and softly quantifying the degree of factual overlap between them. To refine this algorithm, we introduced a granular mechanism based on TF-IDF weights that adjusts the level of fact extraction based on the involved entity types. In addition to the score, which shows the semantic overall based on tangible facts, the constituents of the score, including FP and FN, provide interpretable insights of where the model went wrong by fabricating new facts enumerated in FP or ignoring necessary facts enumerated in FN.

This work represents a step toward enabling generative models to produce more fact-aware outputs. As a next step, we aim to integrate our scoring method into training pipelines, allowing models to better adhere to factual content. One potential direction is to incorporate the score as a soft constraint within the objective function. Alternatively, we may leverage TNs and FPs as negative samples, and TPs as positive samples, and use them to train a contrastive learning algorithm. We plan to further explore these directions in future work.

\section*{Limitations}
While our proposed methodology provides a novel framework for assessing factual consistency, several limitations should be acknowledged. First, our evaluation primarily focused on controlled datasets, and the robustness of the scoring algorithm in highly noisy, real-world clinical or open-domain scenarios remains to be validated. Second, although the TF-IDF–based weighting mechanism allows for some granularity, it may not fully capture the nuanced importance of domain-specific entity types, especially in contexts where subtle distinctions carry significant implications (e.g., between similar medical conditions or treatments). Third, the decomposition of text into atomic facts, while useful for consistency measurement, can introduce subjectivity or error depending on the quality of the fact extraction process. Finally, our approach has not yet been integrated into end-to-end generative model training, and thus its impact on actual model outputs and downstream task performance remains to be empirically demonstrated.

\bibliography{custom}

\end{document}